\documentclass{article}

\usepackage{PRIMEarxiv}

\usepackage[utf8]{inputenc} 
\usepackage[T1]{fontenc}    
\usepackage{hyperref}       
\usepackage{url}            
\usepackage{booktabs}       
\usepackage{amsfonts}       
\usepackage{nicefrac}       
\usepackage{microtype}      
\usepackage{lipsum}
\usepackage{fancyhdr}       
\usepackage{graphicx}       
\usepackage{comment}
\usepackage{amsmath}%
\usepackage{subfig}
\graphicspath{{media/}}     

\title{Dark Transformer: A Video Transformer for Action Recognition in the Dark}

\author{
  Anwaar Ulhaq \\
  Central Queensland University,  \\
  School of Engineering and Technology\\
  Sydney campus,  Australia\\
  \texttt{\{a.anwaarulhaq@cqu.edu.au} \\
}

\begin{document}
\maketitle

\begin{abstract}
Recognising human actions in adverse lighting conditions presents significant challenges in computer vision, with wide-ranging applications in visual surveillance and nighttime driving. Existing methods tackle action recognition and dark enhancement separately, limiting the potential for end-to-end learning of spatiotemporal representations for video action classification. This paper introduces Dark Transformer, a novel video transformer-based approach for action recognition in low-light environments. Dark Transformer leverages spatiotemporal self-attention mechanisms in cross-domain settings to enhance cross-domain action recognition. By extending video transformers to learn cross-domain knowledge, Dark Transformer achieves state-of-the-art performance on benchmark action recognition datasets, including InFAR, XD145, and ARID. The proposed approach demonstrates significant promise in addressing the challenges of action recognition in adverse lighting conditions, offering practical implications for real-world applications.
\end{abstract}


\section{Introduction}
Recognising human actions across diverse environments presents a challenging task in computer vision. Over the past decade, numerous automated action recognition approaches have been developed, employing various modalities \cite{1,ulhaq2016face,ulhaq2018action,haq2012temporal}. Human action recognition in dark conditions is crucial for applications like airport screening or mask-wearing detection and surveillance, especially during global challenges like the pandemic \cite{ulhaq2020computer}, where accurate action recognition is essential for ensuring public health and safety. Traditional feature-based and machine-learning methods have exhibited limited performance within specific domains, leading researchers to focus on deep learning-based techniques \cite{ulhaq2016action}. These techniques enable end-to-end processing of data for simultaneous feature extraction and classification. Convolutional Neural Network (CNN)-based methods have emerged as a prominent approach, leveraging hierarchical feature extraction. Initial layers capture local features, while subsequent layers extract global features. This paper provides an overview of the advancements in deep learning-based action recognition techniques, emphasising the hierarchical feature extraction capabilities of CNNs.


\begin{figure}
\centering
\includegraphics[width=130 mm]{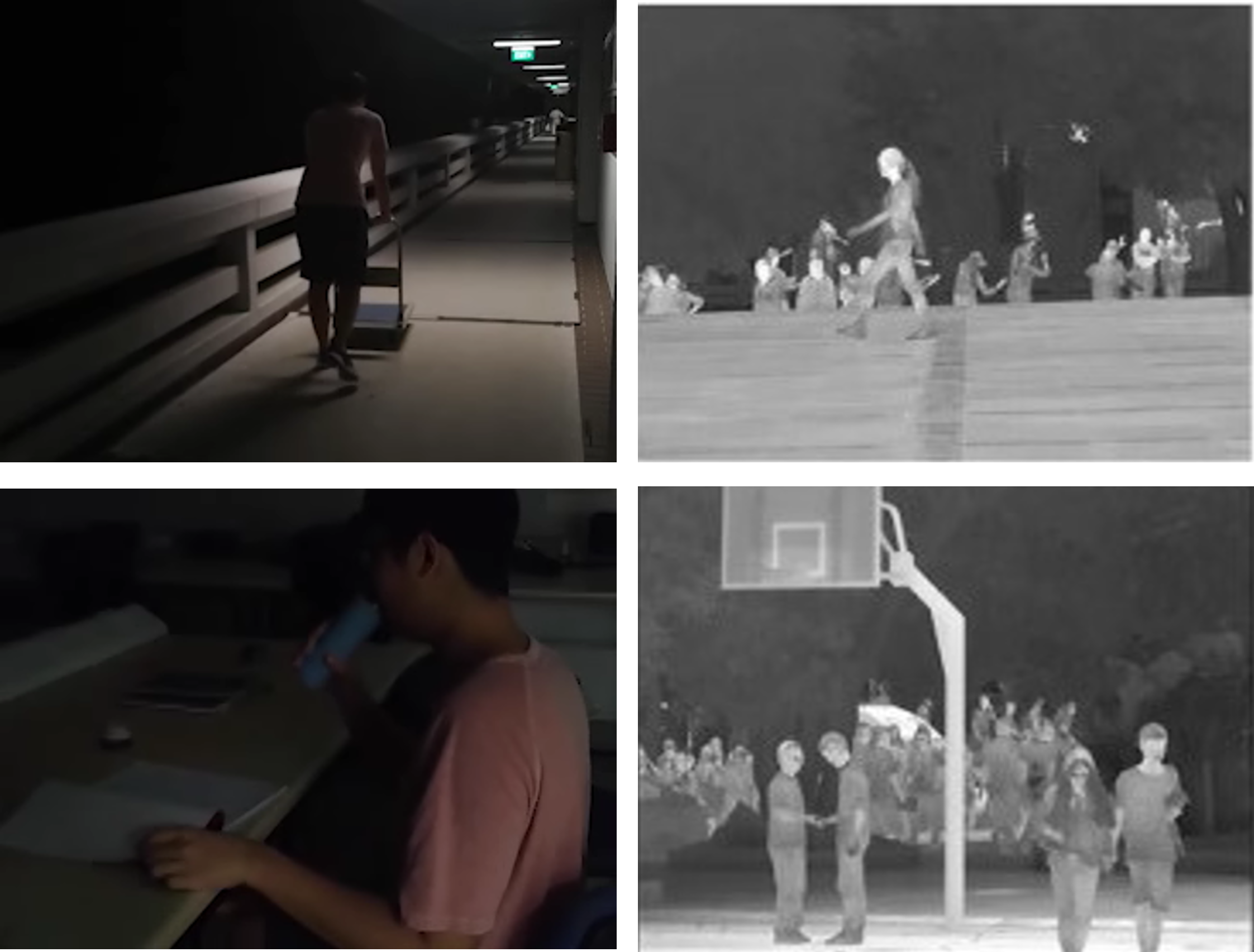} 
\caption{
Challenges in Automated Action Recognition: Capturing Examples of Actions (Push, Walk, Handshaking, Drink) in the Dark from RGB and Infrared Domains }
\label{fig:ACCV22}
\end{figure}

Most existing approaches for action recognition focus on daytime videos or scenarios with adequate lighting conditions, overlooking the challenges of adverse lighting conditions or nighttime action recognition \cite{gondal2011action}. The limited research in the field of action recognition in dark videos can be attributed to two key factors: (i) ineffective data enhancement methods, where visually enhancing dark video frames may not consistently improve action recognition accuracy, and (ii) the scarcity of suitable datasets for exploring action recognition in low-light environments, in contrast to the availability of large action datasets like Kinetics-400 and Kinetics-600 for general action recognition. This paper highlights the need to address action recognition in dark videos and discusses the underlying reasons for the limited research in this area.

Despite their remarkable success in various application scenarios, the generalisation performance of deep neural networks often suffers when encountering new domains due to the domain shift problem. Unsupervised Domain Adaptation (UDA) aims to overcome this challenge by transferring knowledge from a labelled source domain to an unlabeled target domain. As documented in \cite{2,3,4,cdtrans, TVT,haq2010automated}, extensive research efforts are dedicated to addressing the Unsupervised Domain Adaptation problem and alleviating the need for laborious annotations. These endeavours strive to enhance the adaptability and robustness of deep neural networks across diverse domains.

The advent of vision Transformers \cite{Visiontransform} has highlighted the effectiveness of cross-attention mechanisms in aligning diverse distributions, even across different modalities. However, while these techniques have primarily focused on image-based tasks, exploring unsupervised domain adaptation in the video domain, specifically for action recognition, remains relatively unexplored. Consequently, the question arises: How can a similar model be constructed for human action recognition? This paper addresses this gap by investigating the application of cross-attention mechanisms in unsupervised domain adaptation for action recognition tasks in video data.

The Transformer architecture is a suitable model because it explicitly builds contextual support for its representations through self-attention. This architecture has been tremendously effective for sequence modelling tasks compared to traditional recurrent models. It has also achieved remarkable success with visual tasks, and various architectures are being devised for image classification, segmentation, object detection, and video classification. Similarly, cross-domain transformers have been devised for image classification. However, there is still a research gap about cross-domain video transformers \cite{ulhaq2022vision}.

Our work is motivated by (I) the success of vision transformers for visual tasks compared to their CNN counterparts and (II) Domain adaptation methodologies that obtain promising results from transfer learning across domains. In conclusion, our contributions are of three types:

1- We propose Dark Transformer, a weight-sharing triple-branch video transformer framework, for day-night action recognition by extending the Timesformer architecture by integrating unsupervised domain adaptation and leveraging knowledge distillation. 
2—To the best of our knowledge, the proposed model is the first of its kind to use a domain-invariant video transformer architecture for action recognition in the dark. 
3- It outperforms the state-of-the-art on benchmark datasets by a significant margin.

\subsection{Related Work}

The current optical flow estimation methods \cite{8} fail to accurately capture optical flow for action recognition due to the poor quality of dark images. As a result, the performance of two-stream methods \cite{6,7} is also subpar in such conditions. Consequently, image enhancement is often considered a prerequisite for action recognition tasks. One approach, called Multi-scale Retinex \cite{9}, enhances dark images by combining multiple Single Scale Retinex (SSR) outputs into a single output image, providing colour constancy and dynamic range compression. Another approach, LIME \cite{LIME}, improves dark images by estimating and refining the illumination map. However, integrating these approaches into an end-to-end network poses challenges.

Alternatively, another approach is to fuse multiple spectrums to address this issue. Previous works \cite{11,12,ulhaq2016face} have proposed methods for simultaneous action recognition from multiple video streams. Recently, Xu et al. \cite{ARID} introduced the ARID dataset, which focuses on human actions in dark videos and reveals the ineffectiveness of current action recognition models and frame enhancement techniques in dark conditions. Dark-Light networks \cite{Darklight,ulhaq2016face} leverage dark videos and their brightened counterparts, employing a dual-pathway structure for improved video representation and significantly enhancing action recognition performance. However, this approach is computationally intensive.

Inspired by the success of NLP transformers \cite{transformer} and vision transformers \cite{Visiontransform,Swin}, recent advancements have introduced video transformers \cite{Visiontransform,vivit,videoswin}, which achieve state-of-the-art performance in action recognition tasks. In this work, we extend the concept of video transformers to learn cross-domain attentions for action recognition in dark environments, building upon the achievements of previous works.

---------------------- 
\section{The Prpposed Method}
In this Section, we present a comprehensive overview of the architecture of our proposed model. We outline the key components and their interactions within the model to provide a clear understanding of its structure and functioning.

Subsequently, in the subsequent subsections, we delve into the process of video tokenisation and introduce the space-time cross-attention module. We explain how the input videos are tokenised into smaller units to enable efficient processing and information extraction. Additionally, we detail the implementation of the space-time cross-attention module, which facilitates the model's ability to capture both spatial and temporal dependencies in an interconnected manner.

Discussing these aspects, we aim to thoroughly explain the model's design choices and highlight the significance of video tokenisation and space-time cross-attention in achieving robust action recognition performance.

\subsection{Overview of Dark Transformer}

Figures 2 and 3 visually depict the framework of the proposed domain-invariant video transformers (Dark Transformer), illustrating the architectural components and the flow of information within the model. The Dark Transformer framework comprises three weight-sharing transformers, each serving a specific purpose and incorporating different data flows and constraints.

The first branch is called the "Source" branch, which primarily handles the processing and representation learning of the source domain videos. This branch focuses on capturing domain-specific features and patterns from the source domain videos to establish a robust foundation for action recognition.

The second branch, the "Target" branch, handles the target domain videos. This branch specifically deals with the challenges the target domain poses, such as adverse lighting conditions, and adapts the model to recognise actions effectively in such environments. The Target branch utilises the learned knowledge from the Source branch while incorporating domain-specific transformations and adjustments to enhance its performance in the target domain.

Lastly, the "Source-Target" branch integrates information from both the Source and Target branches, facilitating learning domain-invariant representations. This branch aims to align and fuse the learned knowledge from both domains, enabling the model to generalise well across domains and exhibit robustness in action recognition tasks.

The framework's inputs are selected pairs of related video sequences., $X_{1},X_{2}$, where $X_{1}\in D_{source}$, with a given labelled source (daytime) dataset $D_{1}$ and $X_{2}\in D_{target}$, an unlabelled target dataset (nighttime) $D_{2}$, with two different data distributions due to domain shift but they share an identical label space. We aim to train a model that can utilise the valuable source domain knowledge $D_{source}$ to help learn in the target domain $D_{target}$. Each video $X \in R^T \times H \times W \times N$, where N is the number of frames.

As shown in Fig. 3, the source and target videos in the input pair are sent to the source and target branches, respectively. In these two branches, the space-time self-attention module is involved in learning the domain-specific representations. The softmax cross-entropy loss is used to train the action classification. It is worth noting that all three branches share the same classifier due to the same label for two videos.

\begin{figure*}
\centering
\includegraphics[width=100mm]{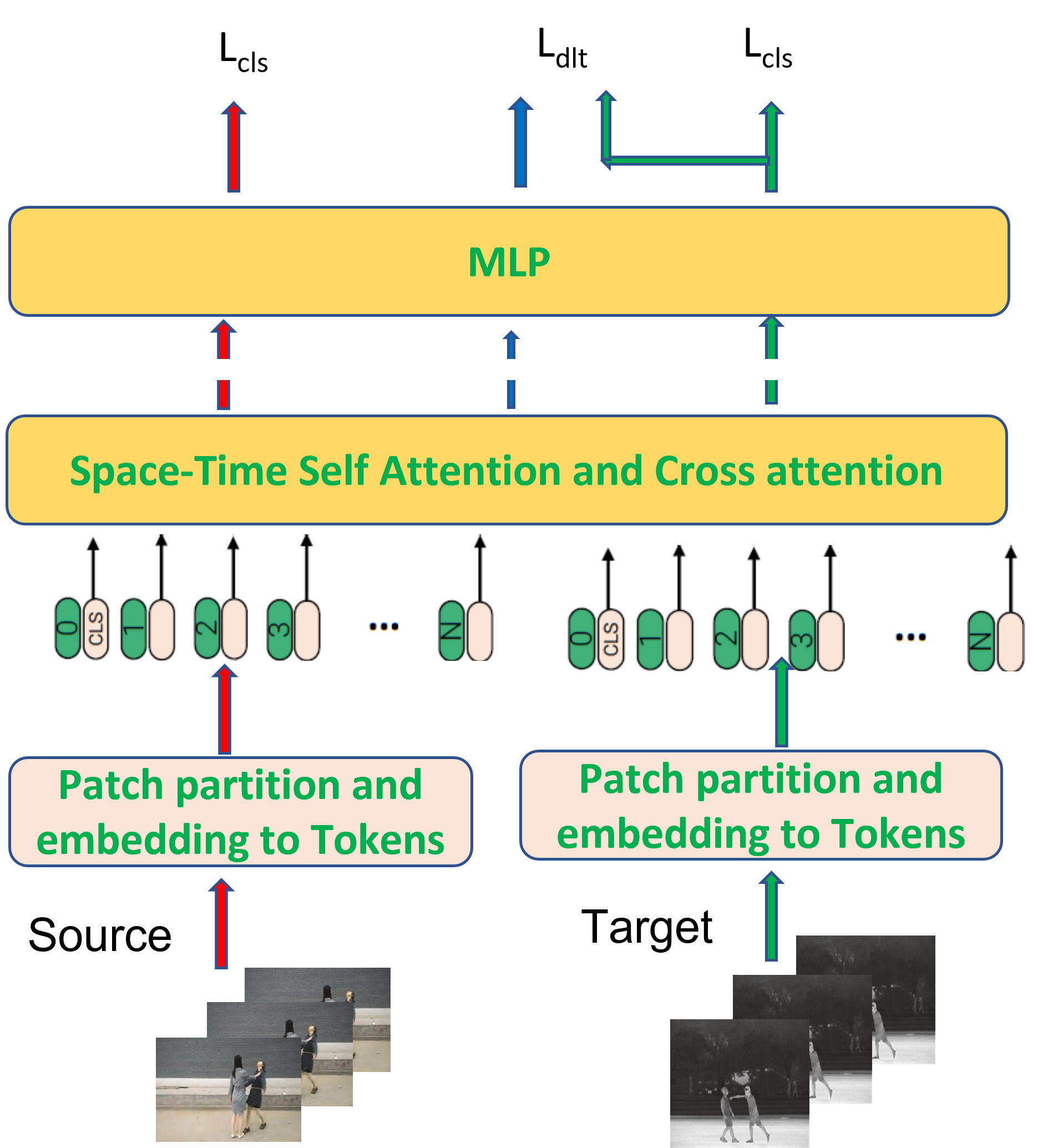} 
\caption{Architectural diagram of the proposed Dark Transformer Transformer during training. The model inputs source (daytime) and target domain (nighttime) videos. However, during inference, only the target domain videos are utilised as a single modality for action recognition.}
\label{fig:ACCV22}
\end{figure*}

\subsection{Embedding Video Clips} 

We employ two straightforward methods to map a video, denoted as X, into a sequence of tokens, denoted as Z. The first method involves tokenising the input video by uniformly sampling frames from the video clip. Each frame is then independently embedded into 2D tokens using the same method as ViT (Vision Transformer). These tokens are concatenated together to form the sequence representation. To handle each frame, we decompose it into M non-overlapping patches, where each patch has a size of S x S. These patches are flattened into vectors, denoted as v(s, t), where s ranges from 1 to M and t ranges from 1 to N, representing the spatial and temporal locations, respectively. Next, each patch is linearly embedded into embedding tokens, denoted as $Z(s, t)$, with additional spatiotemporal position embeddings. These embedding tokens are then passed through the transformer encoder for further processing.

\begin{equation}
Z = (z_{cls}, E_{x_{1}}, \dots, E_{x_{M}}) + P,
\label{equ:dt}
\end{equation},

where the projection by E is equivalent to a 2D convolution, and P is a learned positional embedding. 
\subsection{Space-Time self attentions (S+T) } 

This model forwards all spatiotemporal tokens $Z(s,t)$ extracted from the video through the Dark Transformer transformer encoder. We calculate spatiotemporal self-attention for individual domains for source and target branches separately. The proposed model consists of $L$ encoding blocks. At each block, a query/key/value vector is computed for each patch from the representation $(s;t)$ encoded by the preceding block as follows:

\begin{eqnarray}
Q_{s,t}^{L,h} &=& W_{Q}^{L,h} (Z_{s,t}^{L,h}), \\
K_{s,t}^{L,h} &=& W_{K}^{L,h} (Z_{s,t}^{L,h}), \\
V_{s,t}^{L,h} &=& W_{V}^{L,h} (Z_{s,t}^{L,h}), \quad \text{where } h \in D_{h}
\end{eqnarray}

 Where h is an index over multiple attention heads, and D is the latent dimension. As the direct computation of spatiotemporal self-attention is very expensive, we used a similar strategy to using divided space and time attentions (S+T), where temporal and spatial attention are separately applied one after the other. This method computes temporal attention by comparing each patch with all the patches at the same spatial location in the other frames. Instead of being passed to the MLP, the encoding resulting from applying temporal attention is fed back for spatial attention computation.
 
\begin{equation}
A_{Time} = \text{softmax}\left(\frac{Q_{s,t}^{L,h^{T}}}{\sqrt{D_{h}}} \sum_{t'=1}^{N} K_{0,0}^{L,h}(K_{s,t'}^{L,h})\right),
\label{equ:dt}
\end{equation}

while spatial attention is given as:

 \begin{equation}
A_{Space} = \text{softmax}\left(\frac{Q_{s,t}^{L,h^{T}}}{\sqrt{D_{h}}} \sum_{s'=1}^{N} K_{0,0}^{L,h}(K_{s',t'}^{L,h})\right),
\label{equ:dt}
\end{equation}

 \subsection{Space-Time Cross attentions }
 
The module for cross-attention is imported into the source-target branch. The source-target branch receives inputs from the other two branches. In the $l$-th layer, the query of the space-time cross-attention module comes from the query in the $l$-th layer of the source branch, while the keys and values are from those of the target branch. However, for space and time cross attention, rather than cross each attention one by one, only final attentions from S-T self-attentions are used to reduce computation.  

 \begin{equation}
  A_{Cross (X_{source}, X_{target})} = softmax(\frac{Q_{source}K_{target}^{T}}{sqrt{D_{h}}}.V_{target})
  \label{equ:dt}
\end{equation}

The output of the cross-attention module is then added to the previous layer's output.
.
 \begin{equation}
  Z_{source \rightarrow{target}}^{L} = MCA(LN(Z_{source}^{L-1},Z_{target}^{L-1} )) + Z_{source \rightarrow{target}}^{L-1}
  \label{equ:dt}
\end{equation}

 \begin{figure*}
\centering
\includegraphics[width=120mm]{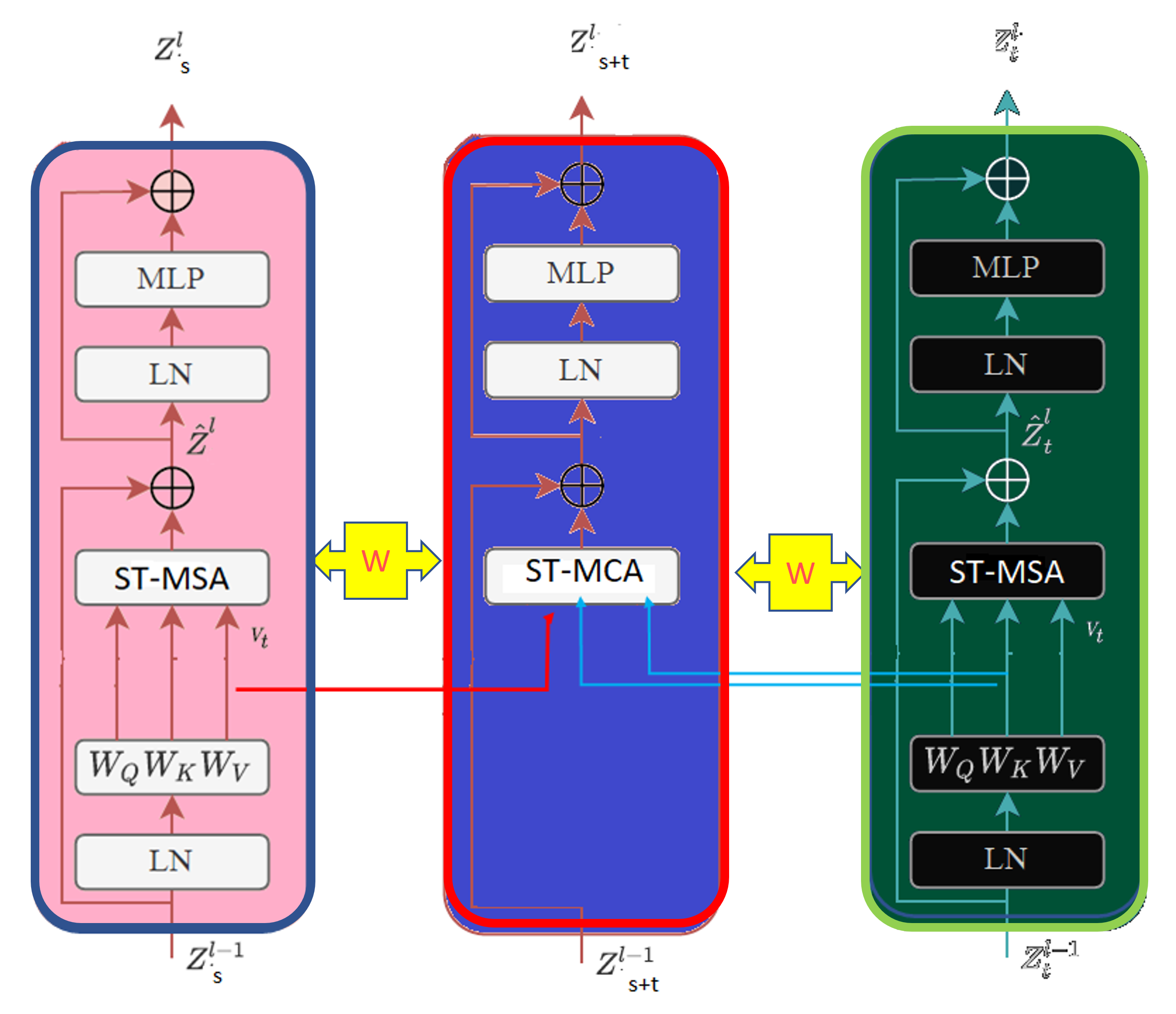} 
\caption{
The architectural diagram showcases the three weight-sharing branches of the Dark Transformer. The middle branch, highlighted in this diagram, plays a crucial role by providing accurate alignment through space-time cross-domain attention. }
\label{fig:ACCV22}
\end{figure*}

The cross-attention module makes the source-target branch features not only align distributions of two domains but also robust to noise in the input pairs. Thus, we utilise the output of the source-target branch as a training guide for the target branch.

 \subsection{Knowledge Distillation }

Both the source-target and target branches are stated, respectively, as teacher and student. We regard the classifier's probability distribution in the source-target branch as a soft label that can be used to supervise the target branch further using a distillation loss.

\begin{equation}
L_{dtl} = \sum_{i}p_{k}log q_{i}
  \label{equ:dt}
\end{equation}

where $q_{i}$ and $p_{i}$  are the probabilities of category i from the source-target and target branches, respectively. 

 It has been demonstrated that ViT \cite{Visiontransform} is only effective when trained on massive datasets. Nonetheless, even the largest video datasets, including Kinetics [35], have several orders of magnitude fewer labelled examples when compared to their image counterparts. As a result, training large models from scratch to achieve high accuracy is an arduous task. Our model uses a pre-trained Timesformer for each domain.

\section{Experimental Results} 

textbfDatasets. This study utilises two datasets in the visible-to-infrared action recognition task: the InFAR dataset and the XD145 dataset. The XD145 dataset, which will be made available at https://sites.google.com/site/yangliuxdu/, serves as the source domain, while the InFAR dataset is employed as the target domain. These datasets provide valuable resources for training and evaluating the proposed approach, allowing for a comprehensive analysis of the model's performance in cross-domain action recognition tasks from visible to infrared domains.

(A) InfAR
The InfAR dataset \cite{Infar} includes 600 video sequences captured by infrared thermal imaging cameras. As depicted in Figure 4, the dataset consists of fight, handclapping, handshake, hug, jog, jump, punch, push, skip, walk, wave 1 (one-hand wave), and wave 2 (two-hand wave), where each action class contains 50 video clips with an average duration of 4 seconds.

The frame rate is 25 frames per second, and the resolution is 293 by 256 pixels. Each video depicts single or multiple actions carried out by single or multiple individuals. Some involve interactions between multiple individuals, as illustrated in Figure 4.

(B) XD145
A visible light action dataset named XD145 reflects the above inFar dataset. In correspondence with the target domain action categories, the XD145 and the InfAR dataset have the same action categories, as shown in Figure 4. The XD145 action dataset consists of 600 video sequences captured by visible light cameras, and there are 50 video clips for each action class. All actions were performed by 30 different volunteers. Each clip lasts for 5 seconds on average. The frame rate is 25 fps, and the resolution is 320 × 240. As shown in Figure 4, the background, pose, and viewpoint variations are considered when constructing the dataset to make our dataset more representative of real-world scenarios.

C) ARID: We also conduct experiments on the first benchmark datasets for action recognition in the dark: ARID \cite{ARID}, which contains 3,784 video clips in 11 action categories with lower brightness and contrast than other AR video datasets. Three splits' Top-1 and Top-5 accuracy averages are reported.

\begin{figure}
\centering
\includegraphics[width=100mm]{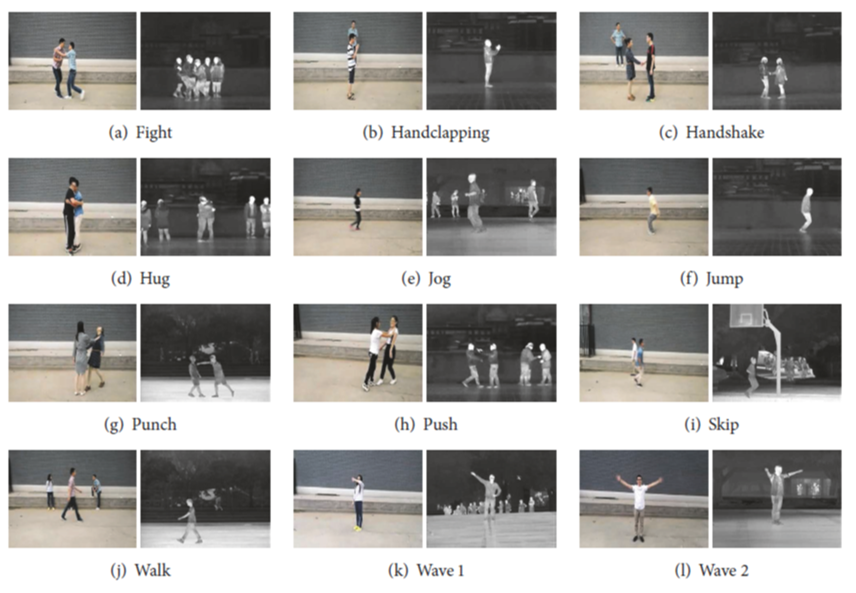} 
\caption{
Examples of visible and infrared actions. Each subfigure displays a visible image (left) from the XD145 dataset's video sequences and an infrared image (right) from the InfAR dataset's video sequences.}
\label{fig:ACCV22}
\end{figure}

\textbf{Inference } In our approach, the input to the network consists of a 32-frame video clip with a stride of 2. This means that consecutive video frames are sampled with a temporal distance of 2 frames, allowing for efficient processing while maintaining temporal context.

We follow a standard procedure during the inference phase to process the video clips from the target domain. This involves applying our trained network to the target domain video, analysing each frame or video clip individually, and aggregating the results to obtain the final prediction or output. This procedure ensures that the network can adapt and generalise well to unseen videos in the target domain, enhancing our approach's overall performance and accuracy in action recognition tasks.

\textbf{Comparisons With State-of-The-Art Methods}

From the analysis of Tables 1 and 2, we observe that different CNN-based feature extractors can effectively extract discriminative features for action recognition tasks. Additionally, Dark Transformer demonstrates a clear improvement in performance for both the InfAR and ARID action datasets.

Comparing our approach to the I3D-Two-stream network, we notice a substantial increase in Top-1 accuracy by 24.73 \%. This finding validates our proposed method's effectiveness and suggests that optical flow may not be as useful for action recognition in dark environments. The performance gain achieved by Dark Transformer highlights its powerful capabilities in handling challenging lighting conditions.

Furthermore, when comparing Dark Transformer to the 3D-ResNet-18 and 3D-ResNet-101 architectures, we observe that deeper network layers lead to higher performance gains. This finding reinforces the notion that deeper networks have the potential to capture more intricate and abstract features, which can significantly benefit action recognition tasks. Dark Transformer, with its enhanced model architecture, achieves even better performance than these deeper network variants.

\begin{figure}
\centering
\includegraphics[width=120mm]{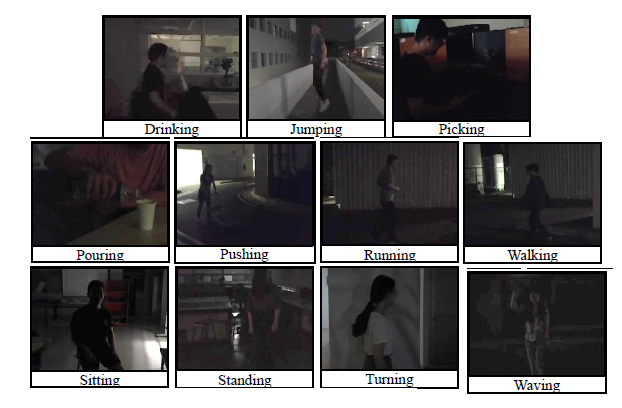} 
\caption{Sample frames illustrating each of the 11 action classes from the ARID dataset.}
\label{fig:ACCV22}
\end{figure}

Similarly, for the InfAR dataset, Dark Transformer demonstrates a remarkable improvement of 6.83 \% in performance compared to the 3D-ResNet-101 architecture. This notable enhancement further emphasises the effectiveness of our approach in tackling action recognition in adverse lighting conditions.

Overall, the experimental results highlight Dark Transformer's superiority to other feature extraction methods and network architectures, establishing it as a highly promising and powerful approach for action recognition in challenging lighting scenarios.

\begin{figure}
\centering
\includegraphics[width=80mm]{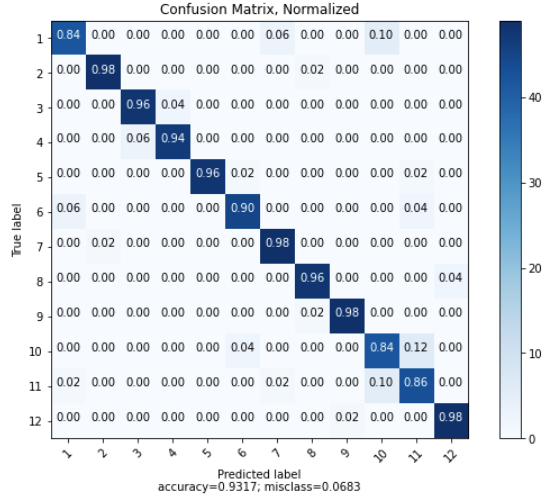} 
\caption{
Confusion matrix for the InfaR action recognition dataset.}
\label{fig:ACCV22}
\end{figure}

\setlength{\tabcolsep}{4pt}
\begin{table}
\begin{center}
\caption{
The Top-1 and Top-5 accuracy results of a few competitive models and ours on the ARID dataset.
}
\label{table:headings}
\begin{tabular}{lll}
\hline\noalign{\smallskip}
Method $\qquad\qquad$& Top-1 & Top-5\\
\noalign{\smallskip}
\hline
\noalign{\smallskip}
VGG(Two-stream) \cite{VGGTwo-stream} & 32.08 & 90.76\\
C3D \cite{C3D} & 40.34 & 94.17\\
3D-SqueezeNet \cite{3DSqueezNet}& 50.15 & 94.16\\
R(2+1)D \cite{Rtwoplusone}& 62.87 & 96.64\\
I3D-Two-stream \cite{I3D}& 73.40 & 97.20\\
3D-ResNext-101\cite{3DResnNet101} & 86.36 &99.52\\
DarkLight-ResNext-101 \cite{2021darklight}& 87.27 &99.47\\
DarkLight-R(2+1)D-34\cite{2021darklight} & 94.04 &99.87\\
Dark Transformer & 98.13 & 99.97\\

\hline
\end{tabular}
\end{center}
\end{table}
\setlength{\tabcolsep}{1.4pt}

\subsection{Ablation study} 

To gain a deeper understanding of the impact of various components in the network, we conducted an ablation study where we systematically removed each component and observed its effect on the performance of Dark Transformer. We explored two different settings for this purpose. Firstly, we varied the number of samples from the target domain during training and evaluated the performance while keeping both space-time self-attentions intact. Secondly, we examined the performance when only time self-attentions were used in one setting and only space self-attentions were used in another. The results are presented in Figure 7, clearly demonstrating the contribution of space-time self-attention to the overall performance. Additionally, the results indicate that temporal attention plays a more crucial role than spatial attention in action recognition tasks.

\setlength{\tabcolsep}{4pt}
\begin{table}
\begin{center}
\caption{
Average recognition accuracy results of a few competitive models and ours on the InfaR dataset.
}
\label{table:headings}
\begin{tabular}{lll}
\hline\noalign{\smallskip}
Method $\qquad\qquad$& Top-1 \\
\noalign{\smallskip}
\hline
\noalign{\smallskip}
iDT \cite{iDT} & 71.35 \\
2 Stream 2D CNN \cite{Infar}& 76.66 \\
2 Stream 3D CNN \cite{C3D}& 77.5 \\
CDFAG \cite{Infar}& 78.55 \\
4-Stream CNN \cite{4} & 83.40 \\
3D-ResNext-101 \cite{3DResnNet101}& 86.36 \\
SCA \cite{CSA}& 84.25\\
Dark Transformer & 95.17 \\
\hline
\end{tabular}
\end{center}
\end{table}
\setlength{\tabcolsep}{1.4pt}

\textbf{Varying the number of input frames} In our previous experiments, we fixed the number of input frames at 32. Now, we increase and decrease the number of frames input to the model, thereby proportionally increasing the number of tokens. We observed
a slight increase in performance  (0.5 $\%$) for an increase in the number of frames to 64. However, a decrease in the number of frames adversely impacts the performance (5 $\%$) if the number of frames is reduced to 16.

Feature visualisation illustrates the performance of the domain-invariant nature of learned features. Figure 8 illustrates that our model's feature visualisation remains almost invariant for both domains.

\begin{figure}
\centering
\includegraphics[width=80mm]{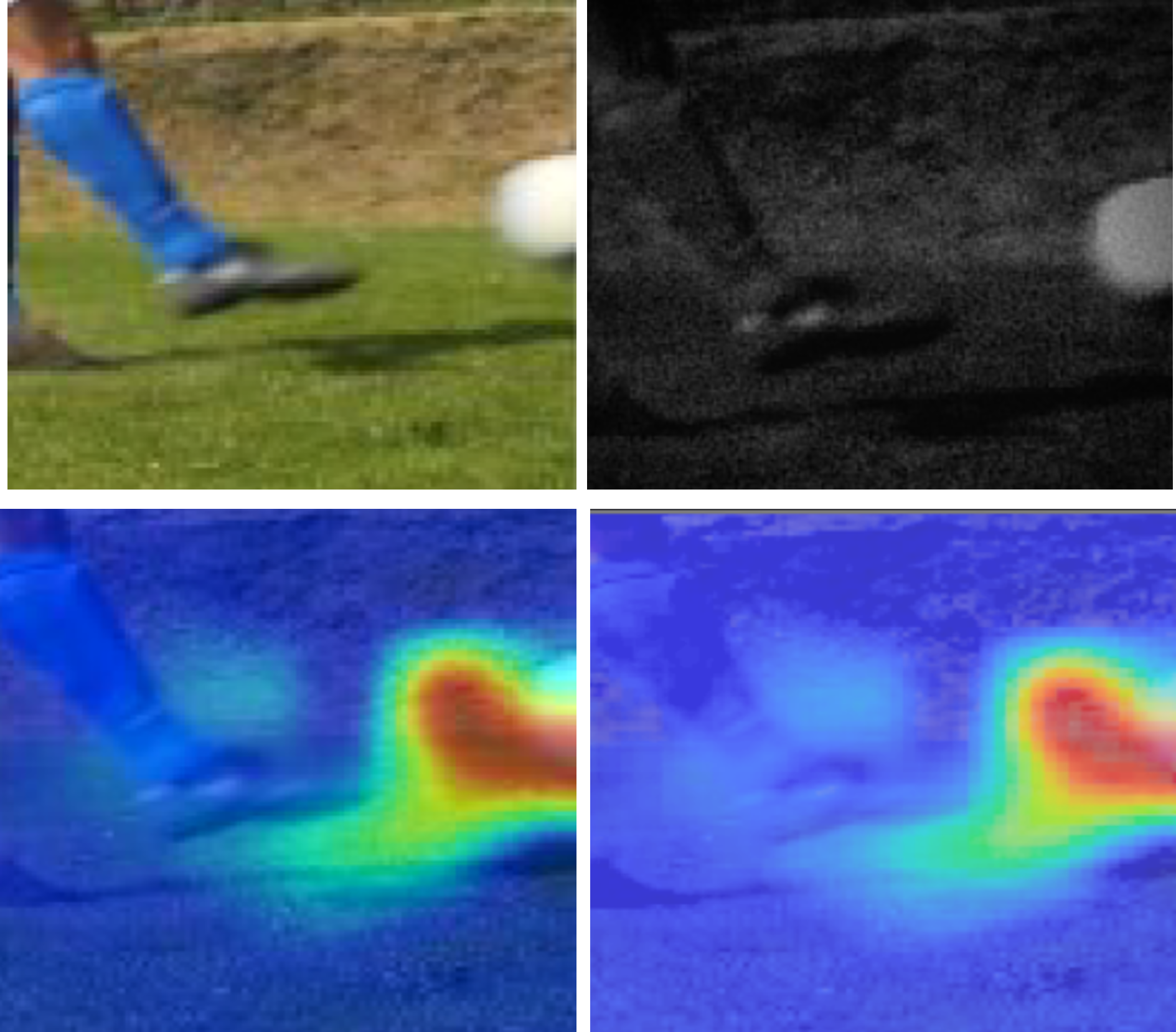} 
\caption{
Attention visualisation shows the domain-in-variance nature of learning by the proposed model, as it shows the similarity between activations.  }
\label{fig:ACCV22}
\end{figure}

The comparisons illustrate that the Dark Transformer network is far better than many other excellent models based on 3D-CNN or two-stream in Top-1 accuracy. We further perform ablation experiments on Dark Transformer to explore the role of each part. It illustrates that space-time self-attention blocks can catch more important spatiotemporal features for AR. In conclusion, applying a video transformer with space-time self-attention with cross-domain learning can achieve the best result.

\begin{figure}
\centering
\includegraphics[width=100mm]{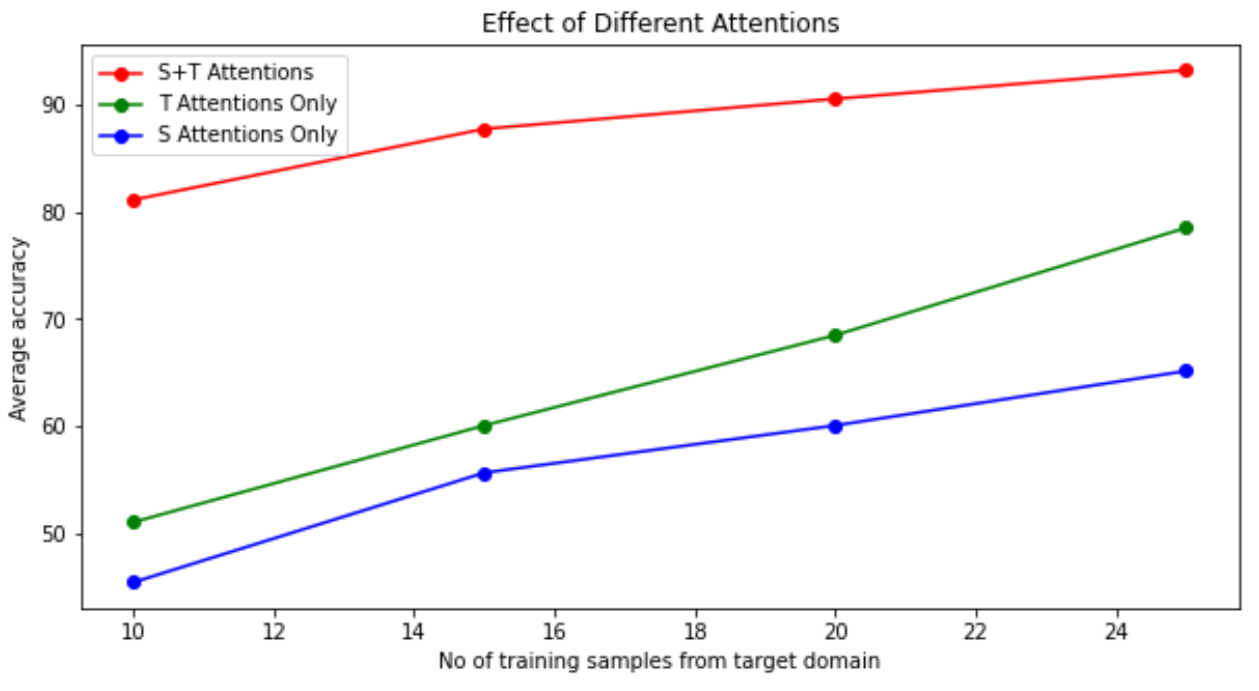} 
\caption{
Ablation study by removing different self-attention heads in the model and observing their impact on performance..}
\label{fig:ACCV22}
\end{figure}

\section{Conclusions}
In conclusion, this paper presents Dark Transformer, a novel video transformer-based model designed to recognise human actions in challenging lighting conditions. By leveraging spatiotemporal self-attention in cross-domain settings, the model learns robust representations that support cross-domain action recognition. Dark Transformer extends the capability of video transformers by incorporating cross-domain knowledge and achieves state-of-the-art results on benchmark datasets, including InFAR, XD145 and ARID, which are known for adverse lighting and diverse action categories. Extensive experiments show that Dark Transformer consistently outperforms existing approaches and establishes a new benchmark for action recognition in low-light environments. These findings highlight the effectiveness of video transformer-based methods in handling illumination variability and point toward promising directions for future research and real-world deployment in visual surveillance and autonomous systems.



\end{document}